\pdfoutput=1
\documentclass{article}
\PassOptionsToPackage{numbers,sort&compress}{natbib}
\usepackage[main, final]{neurips_2026_vericode}
\workshoptitle{VeriCodeGen: AI for Verifiable Coding}
\usepackage[utf8]{inputenc}
\usepackage[T1]{fontenc}
\usepackage{amsmath,amssymb,booktabs,tabularx,array,microtype,xcolor}
\usepackage{newunicodechar} 
\newunicodechar{ℝ}{\ensuremath{\mathbb{R}}}
\newunicodechar{ℕ}{\ensuremath{\mathbb{N}}}
\newunicodechar{⊢}{\ensuremath{\vdash}}
\newunicodechar{∃}{\ensuremath{\exists}}
\newunicodechar{∧}{\ensuremath{\wedge}}
\newunicodechar{≤}{\ensuremath{\le}}
\newunicodechar{≥}{\ensuremath{\ge}}
\newunicodechar{↑}{\ensuremath{\uparrow}}
\newunicodechar{⟨}{\ensuremath{\langle}}
\newunicodechar{⟩}{\ensuremath{\rangle}}
\newunicodechar{₀}{\ensuremath{{}_0}}
\newunicodechar{₁}{\ensuremath{{}_1}}
\newunicodechar{✝}{\ensuremath{\dagger}}
\newunicodechar{⁻}{\textsuperscript{-}}
\newunicodechar{¹}{\textsuperscript{1}}
\usepackage{flafter}
\usepackage{graphicx}
\usepackage{needspace,fvextra}
\definecolor{leanKeyword}{HTML}{173F70}
\definecolor{leanGoal}{HTML}{14544D}
\definecolor{leanError}{HTML}{9A1930}
\definecolor{leanDelimiter}{HTML}{51575D}
\newcommand{\LeanKeyword}[1]{\textcolor{leanKeyword}{#1}}
\newcommand{\LeanGoal}[1]{\textcolor{leanGoal}{#1}}
\newcommand{\LeanError}[1]{\textcolor{leanError}{#1}}
\newcommand{\LeanDelimiter}[1]{\textcolor{leanDelimiter}{#1}}
\usepackage{url}
\usepackage{hyperref}
\definecolor{linkblue}{rgb}{0,0.08,0.45}
\hypersetup{colorlinks=true,linkcolor=black,citecolor=black,filecolor=black,urlcolor=linkblue}
\hypersetup{pdftitle={Learning Lean Feedback for Efficient Proof-Candidate Evaluation on GPUs},pdfauthor={Lazar Milikic, Etienne Bamas, Emmanuel Abbe, Viktor Kun\v{c}ak},pdfsubject={NeurIPS 2026 Workshop on AI for Verifiable Coding (VeriCodeGen)}}
\newcommand{\NA}{\textemdash}
\newcommand{\hybridcutoff}{90}
\newcommand{\hybridevalmean}{90}
\newcolumntype{Y}{>{\raggedright\arraybackslash}X}
\title{Learning Lean Feedback for\\Efficient Proof-Candidate Evaluation on GPUs}
\author{%
  Lazar Milikic \qquad Etienne Bamas \qquad Emmanuel Abbe \qquad Viktor Kun\v{c}ak \\
  EPFL, Lausanne, Switzerland \\
  \texttt{\{lazar.milikic,etienne.bamas,emmanuel.abbe,viktor.kuncak\}@epfl.ch}
}
\begin{document}
\maketitle

\begin{abstract}
Training language models for formal theorem proving requires large numbers of verifier calls. Slow checks and timeouts often leave GPUs waiting and withhold rewards from potentially correct proofs. We train neural evaluators to predict Lean~4 verdicts and compiler feedback on GPUs using 4.84M labelled programs. We compare two validity classifiers with a feedback model that annotates proof attempts with predicted goal states and error messages. A two-stage evaluator that screens candidates with a classifier and scores its positive predictions with the feedback model reaches 87.8\% accuracy in predicting whether Lean accepts a candidate proof, evaluated on a 68{,}932-program benchmark collected from public provers. The combined evaluator processes 3.5 candidates per second on one H100, 4.2$\times$ the throughput of our 64-worker Lean baseline with environment reuse. Finally, we combine our evaluator with Lean verdicts to obtain a hybrid reward system that shortens GRPO training steps by approximately 31\%. After one GRPO epoch with hybrid rewards, the prover achieves nearly the same miniF2F pass@1 as with Lean-only rewards (62.2\% versus 62.4\%) and a pass@32 of 72.4\% versus~71.3\%.
\end{abstract}

\section{Introduction}
\label{sec:intro}
Proof assistants provide automatically checked training signals for language models that generate formal proofs \citep{demoura2021lean,mathlib2019}. Lean-based provers use these signals for supervised learning, proof search, and reinforcement learning \citep{lin2025goedel,lin2025goedelv2,ren2025deepseekproverv2,wang2025kimina,chen2025seedprover,hubert2026olympiad}. Checking candidates at scale creates a substantial computational workload. Existing systems reduce this overhead through parallel workers, reuse of loaded environments, and compiler optimizations \citep{dossantos2025kimina,hubert2026olympiad}.

Even with these optimizations, slow checks can delay reward computation for an entire training batch. In our 8B prover's Group Relative Policy Optimization (GRPO) training \citep{shao2024deepseekmath} using the efficient verl framework \citep{sheng2024hybridflow}, a 64-worker Lean pool takes about 420\,s to score 1{,}024 candidate proofs with a 60\,s timeout per proof, about 40\% of a 1{,}050\,s training step. Timeouts also withhold rewards from some correct proofs: in an audit with a larger verification budget, 58.1\% of the resolved timeouts are valid (Section~\ref{sec:grpo}). A separate comparison with Kimina Lean Server finds similar verification times on 10{,}000 proofs, providing an external check on our pipeline's efficiency (Appendix~\ref{app:kimina_comparison}).

We study the trade-off between evaluation speed and verdict accuracy when predicting Lean's response to a candidate proof. We train two validity classifiers and a feedback model that annotates a supplied program with predicted goal states and error messages. Combining a classifier with the feedback model gives a two-stage evaluator, or \emph{cascade}, that predicts a candidate's validity on the GPU. These predictions can supplement Lean when exact checks are slow.

\begin{figure}[ht]
\centering
\includegraphics[width=\linewidth]{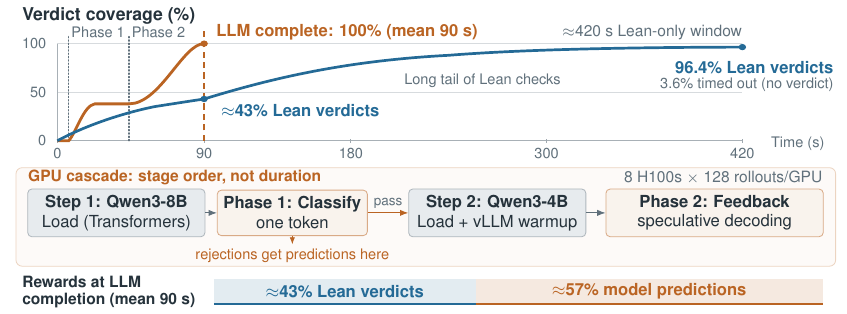}
\caption{\textbf{Hybrid rewards from concurrent Lean and GPU evaluation.} Each batch contains 1{,}024 candidate proofs, checked by 64 Lean workers with a 60\,s timeout per proof. Lean-only batch scoring takes approximately 420\,s on average. Hybrid scoring ends when GPU evaluation finishes (\hybridevalmean\,s on average), retaining completed Lean verdicts and using model predictions for unfinished checks, which are cancelled. The Lean curve beyond this point shows the Lean-only reference. The 3.6\% timeout share is measured over the 24{,}359 candidates with extractable code in the replay cohort (Appendix~\ref{app:grpo}). Curves and phase durations are schematic; flat neural regions represent loading and warmup.}
\label{fig:grpo-hybrid}
\end{figure}

Our hybrid reward protocol runs Lean and the learned evaluator concurrently on each batch (Figure~\ref{fig:grpo-hybrid}). When the evaluator finishes, after 90\,s on average, we retain the Lean verdicts already returned and use model predictions for the remaining candidates. To test whether the learned predictions are useful, we also train with Lean alone under a 13\,s per-proof timeout. This matches the hybrid protocol's average batch scoring time of 90\,s but yields lower miniF2F pass@1 and pass@32 after one epoch of GRPO.

\Needspace{4\baselineskip}
We make three contributions:
\begin{enumerate}
\item \textbf{Fast GPU evaluation with a measured accuracy--cost trade-off.} Our cascade reaches 87.8\% accuracy on 68{,}932 candidates from three public provers. With speculative decoding, it processes 3.5 candidates per second on one NVIDIA HGX H100-SXM5-80GB GPU, 4.2$\times$ the throughput of our 64-worker Lean reference.
\item \textbf{Hybrid rewards for GRPO training.} Hybrid rewards shorten training steps by approximately 31\%, retain essentially the same miniF2F pass@1 as standard Lean-only rewards, and largely preserve the initial prover's pass@32. At the same scoring time, the 13\,s Lean-only baseline performs worse on both metrics.
\item \textbf{A dataset and benchmark for learned proof evaluation.} We construct a 4.84M-program compiler-labelled training corpus and a 68{,}932-program benchmark from three public provers on miniF2F and ProofNet. We release the datasets and model checkpoints on Hugging Face and the project repository on GitHub.\footnote{Datasets and models: \href{https://huggingface.co/collections/EPFL-LARA/learning-lean-feedback}{\texttt{huggingface.co/collections/EPFL-LARA/learning-lean-feedback}}\\\hspace*{1.8em}Project repository: \href{https://github.com/epfl-lara/learning-lean-feedback}{\texttt{github.com/epfl-lara/learning-lean-feedback}}}
\end{enumerate}

\section{Related work}
\label{sec:related}
\paragraph{Learning from proof-assistant feedback.}
Existing provers use proof-assistant feedback to learn, search, and repair proofs. DeepSeek-Prover and Goedel-Prover combine generated data with reinforcement learning and self-correction \citep{xin2024deepseekprover,lin2025goedel,xin2024deepseekprover15,ren2025deepseekproverv2,lin2025goedelv2}. LeanProgress predicts the number of remaining proof steps to guide search \citep{george2025leanprogress}. Baldur and APRIL learn to repair failed proofs using diagnostics \citep{first2023baldur,wang2026repair}, while Compile to Compress uses compiler outputs for local refinement without retaining a long interaction history \citep{li2026compilecompress}. These works use feedback to improve proof success. Our goal is orthogonal: rather than a new reward signal or search strategy, we aim for comparable results in less time by reducing reliance on CPU-bound verification.

\paragraph{Neural compilation and efficient proof checking.}
Learned feedback prediction complements work on neural compilers and efficient access to exact verifiers. Meta LLM Compiler targets compiler optimization and intermediate representations, while LEGO-Compiler translates high-level programs into assembly \citep{cummins2024llmcompiler,zhang2025legocompiler}. SURGE benchmarks LLM prediction of program execution, including Lean verdicts and error messages, and studies gains from fine-tuning \citep{lyu2025surge}. Our work evaluates specialized Lean models and their use in hybrid GRPO rewards. LeanDojo and LeanInteract expose proof-environment interaction \citep{yang2023leandojo,leaninteract}, and Kimina Lean Server combines parallel Lean processes with import reuse \citep{dossantos2025kimina}. Generative Compilation obtains real compiler feedback on partial programs during generation, demonstrated for Rust \citep{muendler2026generativecompilation}. LeanDojo, LeanInteract, Kimina, and Generative Compilation retain exact checks and are orthogonal to our approach; faster Lean verification could increase exact-check coverage in our hybrid rewards.

\paragraph{Evaluating predicted feedback.}
\citet{ananthula2026decodable} show that models can encode verifier information while generating explanations that do not faithfully describe the computation. We therefore compare predicted feedback directly with Lean's messages and their locations in the source.

\section{Dataset construction}
\label{sec:benchmark}
We use compiler-labelled Lean programs to train the evaluators, select their parameters, and evaluate predictions on public-prover outputs. Table~\ref{tab:composition} summarizes the training corpus, in-house validation data, and evaluation benchmark. The datasets and model checkpoints are publicly available.

\begin{table}[bthp]
\caption{Training, validation, and benchmark data. Benchmark rows combine Pythagoras \citep{ongjunleang2026pythagoras}, Goedel \citep{lin2025goedelv2}, and DeepSeek \citep{ren2025deepseekproverv2} outputs. Valid fractions describe the proof correctness rate over all candidate programs, not theorem-level pass rates.}
\label{tab:composition}
\centering\small
\begin{tabularx}{\linewidth}{@{}Yrr@{}}
\toprule
Data & Programs & Valid fraction \\
\midrule
Training corpus & 4.84M & 44.8\% \\
In-house validation (miniF2F and ProofNet) & 27{,}210 & 38.9\% \\
\midrule
miniF2F benchmark candidates & 41{,}638 & 61.7\% \\
ProofNet benchmark candidates & 27{,}294 & 9.0\% \\
Combined benchmark (deduplicated) & 68{,}932 & 40.8\% \\
\bottomrule
\end{tabularx}
\end{table}

\subsection{Training corpus}
The training corpus combines public Lean proofs with additional proof attempts generated to increase the fraction of invalid examples. We draw theorem statements and proofs from NuminaMath-LEAN \citep{numina2024}, the Goedel-Prover SFT corpus and problem set \citep{lin2025goedelv2}, and APRIL proof-repair traces \citep{wang2026repair}. We fine-tune Qwen3-8B-Base \citep{yang2025qwen3} on valid proofs from this pool, then use it to sample further attempts on the same theorems. Lean v4.15.0, released on January 4, 2025, checks every attempt and supplies its verdict and feedback. We use this version to match NuminaMath-LEAN and to allow comparison with existing work; a newer version would not change the nature of the task (Appendix~\ref{app:audit}). The resulting corpus contains 4.84M programs, of which 44.8\% are valid.

\subsection{Validation data}
We select hyperparameter settings for our learned Lean~4 evaluators using 27{,}210 candidate programs, of which 38.9\% are valid. The candidates are generated from miniF2F and ProofNet statements by the in-house theorem prover, trained on the NuminaMath-LEAN and the Goedel-Prover datasets. These data are separate from the evaluation benchmark and are used to choose decision thresholds and the self-consistency voting rule described in Section~\ref{sec:evaluators}.

\subsection{Evaluation benchmark}
The evaluation benchmark tests predictions on complete Lean programs from three public provers. It contains 68{,}932 deduplicated candidates from Pythagoras-Prover-4B \citep{ongjunleang2026pythagoras}, Goedel-Prover-V2-8B \citep{lin2025goedelv2}, and DeepSeek-Prover-V2-7B \citep{ren2025deepseekproverv2}, each run on miniF2F and ProofNet \citep{zheng2022minif2f,azerbayev2023proofnet}. Each candidate includes its imports and surrounding source context. All labels use Lean v4.15.0, matching evaluator training. The collection pipeline reuses loaded Lean environments across requests to avoid repeatedly loading imports. It runs 64 concurrent verification jobs with incremental compilation caching, using the environment-reuse approach described by Kimina Lean Server \citep{dossantos2025kimina}. All six runs use dual-socket AMD EPYC 7543 CPUs (64 cores), 1{,}008\,GiB RAM, and a 180\,s per-command timeout. Section~\ref{sec:results} reports the measured throughput of this Lean REPL pool.

\paragraph{Validity labels and exclusions.}
Labels record whether a candidate satisfies the collection verifier's acceptance policy. A passing proof is valid; compilation failures and proofs of the wrong statement are invalid. Timeouts, infrastructure exceptions, and candidates rejected before compilation are excluded from the labelled corpus. Accuracy is therefore measured on candidates for which the verifier returned an eligible outcome within its budget. Across 81{,}660 collected candidates (total number, before deduplication), 507 (0.6\%) produce exceptions, including 418 timeouts. Appendix~\ref{app:audit} specifies the pre-checks, forbidden constructs, and expected-theorem matching.

\section{Learned Lean~4 evaluators}
\label{sec:evaluators}
We compare direct validity prediction with generation of compiler feedback, then combine them in a two-stage evaluator called a \emph{cascade}. Before supervised training, all three models receive continued pretraining on approximately 130M tokens of general Lean source code and documentation. They then train on verdicts or annotated programs from the compiler-labelled corpus in Section~\ref{sec:benchmark}. Table~\ref{tab:training} summarizes the models and their training settings.

\begin{table}[tbh]
\caption{Evaluator training. Models first receive continued pretraining on approximately 130M Lean tokens, then one supervised epoch. Context lengths and optimization settings below refer to supervised training.}
\label{tab:training}
\centering\small
\begin{tabularx}{\linewidth}{@{}lYYY@{}}
\toprule
 & Encoder & One-token & Feedback \\
\midrule
Backbone & ModernBERT-large (395M) & Qwen3-8B & Qwen3-4B \\
Pretraining task & Predict masked tokens (30\% masking) & Predict next token & Predict next token \\
Supervised target & Binary validity label & One \texttt{true}/\texttt{false} token & Source annotated with feedback \\
Context (tokens) & 8{,}192 & 32{,}768 & 32{,}768 \\
Learning rate & $2\times10^{-5}$ & $10^{-5}$ & $10^{-5}$ \\
Batch size & 64 & 8 & 8 \\
Update steps & 75{,}251 & 16{,}250 & 55{,}470 \\
\bottomrule
\end{tabularx}
\end{table}

\subsection{Verdict classifiers}
The \emph{encoder}, ModernBERT-large (395M parameters) \citep{warner2025modernbert}, predicts validity in one forward pass. Its context limit is 8{,}192 tokens, so longer programs retain their first 2{,}048 and last 6{,}144 tokens.

The \emph{one-token model}, Qwen3-8B \citep{yang2025qwen3}, is trained to output a single \texttt{true} or \texttt{false} token after the input program. At inference, we normalize the probabilities of these two tokens to obtain a validity score between 0 and 1. Scoring requires one forward pass and no text generation.

\subsection{Feedback prediction and validity}
The \emph{feedback model}, Qwen3-4B, takes an unannotated Lean program and predicts a copy with goal states and error messages inserted as Lean comments. A goal state lists the local assumptions and the proposition still to be proved; a tactic is a command that transforms that state. We call this annotated program a \emph{feedback trace} (Figure~\ref{fig:feedback-example}); the \emph{reference trace} is the annotation obtained by executing Lean.

\begin{figure}[htbp]
{\small\begin{Verbatim}[commandchars=\\\{\}]
\LeanKeyword{theorem} mathd_algebra_440 (x : ℝ) (h₀ : 3 / 2 / 3 = x / 10) : x = 5 := \LeanKeyword{by}
  \LeanDelimiter{/- <feedback>}
  \LeanGoal{-- proof state:}
  \LeanGoal{x : ℝ}
  \LeanGoal{h₀ : 3 / 2 / 3 = x / 10}
  \LeanGoal{⊢ x = 5}
  \LeanDelimiter{</feedback> -/}
  \LeanKeyword{have} h₁ : 3 / 2 / 3 = 1 / 2 := \LeanKeyword{by} \LeanKeyword{norm_num}
  \LeanDelimiter{/- <feedback> ... goal state elided ... </feedback> -/}
  \LeanKeyword{rw} [h₁] \LeanKeyword{at} h₀
  \LeanDelimiter{/- <feedback>}
  \LeanError{-- type: error, msg: tactic 'rewrite' failed, did not find}
  \LeanError{-- instance of the pattern in the target expression}
    \LeanError{3 / 2 / 3}
  \LeanDelimiter{... hypothesis context elided ...}
  \LeanDelimiter{</feedback> -/}
\end{Verbatim}
}
\caption{Predicted feedback trace for an invalid miniF2F proof attempt. The model inserts goal states and an error message as Lean comments; generation stops at the predicted rewrite error. Elisions are marked.}
\label{fig:feedback-example}
\end{figure}

\paragraph{Feedback trace construction.}
The model learns to reproduce the annotations that Lean produces for the input program. We execute each program through Lean's interactive read--eval--print loop (REPL) with tactic tracing enabled. The annotation pipeline inserts goal-state comments before their tactic lines and compiler messages after the corresponding source lines. The comments record compact state updates, omitting unchanged hypotheses and printing new or changed declarations. It stops at the first error, retaining the program and feedback up to that point. Appendix~\ref{app:feedback-construction} details source alignment, hypothesis deduplication, and message formatting.

Figure~\ref{fig:feedback-example} illustrates the model's predicted trace for an invalid miniF2F candidate. Given the original program without annotations, the model first predicts the goal state before the \texttt{have} tactic. It then correctly predicts Lean's rewrite error at \texttt{rw [h₁] at h₀} and stops generation. For a valid proof, the target trace instead reaches the end of the program without an error block. Figures~\ref{fig:feedback-paraphrase} and~\ref{fig:feedback-invalid} in the appendix show complete comparisons of predicted and Lean-generated traces.

\paragraph{Deriving a verdict from predicted feedback.}
We predict invalid if the generated trace contains an error marker, ends with a proof-state block with no following source code, or is cut off at the output-token limit instead of ending with an end-of-sequence (EOS) token. Otherwise, we predict valid. Appendix~\ref{app:audit} gives the parsing details.

\subsection{Improving verdict accuracy}
We investigate two complementary techniques for improving verdict accuracy: a \emph{cascade} that requires agreement between two evaluators, and \emph{self-consistency} that aggregates several predictions from the feedback model. Each can be used on its own, and they can also be combined.

\paragraph{Cascade: agreement between evaluators.}
The \emph{cascade} combines the two Qwen models to reduce incorrect acceptance and avoid generating feedback for every candidate. The one-token model acts as a \emph{gate}: candidates scoring below 0.5 are rejected; the others reach the feedback model. By default, the feedback stage generates one trace with greedy decoding. The cascade accepts a candidate only if both stages accept it. The second stage can therefore remove false positives but cannot recover valid proofs rejected by the gate. We use the greedy cascade for hybrid GRPO rewards: it improves verdict accuracy over the feedback model alone on the benchmark (Table~\ref{tab:accuracy}) and GRPO replay (Appendix~\ref{app:grpo}).

\paragraph{Self-consistency (SC): agreement across feedback traces.}
Self-consistency (SC) uses repeated sampling to make the feedback model's verdict less dependent on a single generated trace. For each candidate, we sample five traces at temperature 0.8, derive a verdict from each using the feedback model's verdict rule, and accept the candidate if at least four predict validity. This technique uses only the feedback model and requires more generation than a single trace. We evaluate it both as a standalone evaluator and in combination with the cascade: candidates that pass the one-token gate are then scored using the four-of-five rule instead of a single feedback trace.

\paragraph{Validation and parameter selection.}
We select evaluator parameters and decision thresholds using the in-house validation data in Section~\ref{sec:benchmark}. This includes the four-of-five self-consistency rule and the calibrated classifier thresholds reported below. All settings are fixed before evaluation on the benchmark outputs from the three public provers.

\section{Accuracy, transfer, and serving cost}
\label{sec:results}
\subsection{Prediction accuracy and candidate ranking}
We evaluate verdict prediction on our 68{,}932-program evaluation benchmark (Section~\ref{sec:benchmark}). With 40.8\% valid candidates, accuracy alone is insufficient to characterize performance. We also report metrics that account for class imbalance or assess ranking. F1 balances precision and recall; macro F1 averages the scores for the valid and invalid classes. Matthews correlation coefficient (MCC) summarizes the confusion matrix on a $-1$ to $1$ scale, with 0 indicating no correlation. Area under the ROC curve (AUROC) measures how often a valid candidate scores above an invalid one, with half credit for ties.

\begin{table}[tbh]
\caption{Verdict prediction on the full evaluation benchmark. Self-consistency (SC) samples five traces at temperature 0.8 and requires four valid votes; all other feedback settings use greedy decoding. $\dagger$: threshold selected on in-house validation data; other classifier and gate thresholds are 0.5. Best values in each metric are bold.}
\label{tab:accuracy}
\centering\small
\begin{tabularx}{\linewidth}{@{}Yrrrr@{}}
\toprule
Evaluator & Accuracy & Macro F1 & MCC & AUROC \\
\midrule
Cascade, SC feedback stage @4/5 & \textbf{0.888} & \textbf{0.887} & \textbf{0.783} & \NA \\
Cascade, calibrated gate (0.82)$^\dagger$ & 0.887 & 0.885 & 0.778 & \NA \\
Cascade: one-token $\to$ feedback & 0.878 & 0.877 & 0.772 & \NA \\
Cascade, FP8 feedback stage & 0.877 & 0.876 & 0.770 & \NA \\
Feedback 4B, 5-sample SC @4/5 & 0.874 & 0.873 & 0.762 & \textbf{0.919} \\
Feedback 4B & 0.851 & 0.851 & 0.733 & \NA \\
One-token 8B, calibrated (0.86)$^\dagger$ & 0.831 & 0.831 & 0.683 & 0.905 \\
One-token 8B & 0.752 & 0.752 & 0.592 & 0.905 \\
ModernBERT-large & 0.586 & 0.563 & 0.372 & 0.862 \\
\bottomrule
\end{tabularx}
\par\smallskip\parbox{\linewidth}{\footnotesize AUROC uses classifier scores or the fraction of valid votes; it is omitted for configurations reporting only a binary decision. Self-consistency takes 4.2$\times$ the inference wall-clock time of greedy feedback generation.}
\end{table}

Combining the two Qwen models improves accuracy over either model alone (Table~\ref{tab:accuracy}). The default cascade with greedy feedback decoding reaches 0.878 accuracy and 0.772 MCC, with a 95\% bootstrap accuracy interval of 0.863--0.892. Threshold calibration improves the one-token baseline but does not close the accuracy gap. Selecting its threshold on the validation set raises test accuracy from 0.752 to 0.831. Matching the cascade's validation-set false-positive rate instead gives 0.836 accuracy. Calibrating the cascade gate raises accuracy to 0.887 while reducing valid recall, the fraction of valid candidates retained, from 0.974 to 0.937. Self-consistency improves accuracy when acceptance requires a strong majority. Sampling five traces at temperature 0.8 and requiring four valid votes raises feedback-model accuracy from 0.851 to 0.874 and cascade accuracy from 0.878 to 0.888. A simple three-of-five majority improves standalone feedback accuracy by only 0.3 percentage points. The four-of-five cascade gain has a paired 95\% bootstrap interval of 0.4--1.7 percentage points (Appendix~\ref{app:deployment}); its additional generation cost is reported below.

\begin{table}[t]
\caption{Accuracy by prover on the evaluation benchmark. Each row combines miniF2F and ProofNet outputs; their proportions and the fraction of valid candidates differ across provers.}
\label{tab:generalization}
\centering\small
\begin{tabularx}{\linewidth}{@{}Yrrrr@{}}
\toprule
Prover & Valid fraction & Feedback & One-token & Cascade \\
\midrule
Goedel-Prover-V2-8B & 0.530 & 0.867 & 0.819 & 0.893 \\
DeepSeek-Prover-V2-7B & 0.408 & 0.843 & 0.765 & 0.878 \\
Pythagoras-Prover-4B & 0.328 & 0.848 & 0.688 & 0.865 \\
\bottomrule
\end{tabularx}
\par\smallskip\parbox{\linewidth}{\footnotesize The one-token model and cascade gate use threshold 0.5 for every prover.}
\end{table}

The cascade has the highest accuracy among the three compared configurations for each public prover (Table~\ref{tab:generalization}). Feedback-model accuracy varies less across provers than one-token accuracy: 0.843--0.867 versus 0.688--0.819. The prover outputs also differ in theorem mix, valid fraction, and proof length, so these results describe transfer across their observed outputs.

\paragraph{Ranking candidates for verification.}
Ranking is especially useful on ProofNet, where valid candidates are rare and rejecting every candidate would already give high accuracy. An offline replay tests how many valid proofs can be recovered with a fixed number of Lean checks. Each candidate pool groups attempts from the same benchmark, prover run, and problem name. We rank predicted-valid candidates first and break ties with the one-token score. On ProofNet, checking one candidate per pool uses 3.7\% of all compilation calls and recovers a valid proof in 57.2\% of pools known to contain one, versus 47.9\% under random ordering. At two candidates per pool, the rates are 72.3\% and 63.2\%. ProofNet reuses some problem names, so a small number of pools combine distinct theorems; these figures describe coverage of the recorded pools. Appendix~\ref{app:deployment} details the replay.

\subsection{Feedback quality on valid and invalid proofs}
\label{sec:feedback}
\begin{table}[htbp]
\caption{Predicted feedback compared with Lean v4.15.0 annotations on 2{,}004 evaluation-benchmark programs, balanced by prover and validity label. Appendix~\ref{app:feedback-metrics} defines all metrics.}
\label{tab:feedback}
\centering\small
\begin{tabularx}{\linewidth}{@{}Yrrr@{}}
\toprule
Metric & Overall & Valid & Invalid \\
\midrule
Block F1, exact & 0.796 & 0.969 & 0.623 \\
Anchored block F1 & 0.795 & 0.969 & 0.621 \\
Token F1 & 0.852 & 0.983 & 0.721 \\
Error-presence agreement & 0.840 & 0.974 & 0.707 \\
\bottomrule
\end{tabularx}
\end{table}
We assess feedback on a subset of the evaluation benchmark containing 2{,}004 programs: 334 for each prover--validity combination, giving 1{,}002 valid and 1{,}002 invalid programs. Table~\ref{tab:feedback} summarizes the metrics defined in Appendix~\ref{app:feedback-metrics}; Appendix~\ref{app:audit} gives the sampling procedure. Block F1 is 0.969 on valid programs and 0.623 on invalid programs, showing much closer agreement with Lean on valid proofs. We hypothesize that rare and varied failure modes make invalid traces harder to predict.

Anchored block F1 additionally requires each matched block to occur at the same source position. Its scores, 0.969 on valid programs and 0.621 on invalid programs, are almost unchanged from block F1, suggesting that misplaced matching blocks explain little of the gap. Both metrics require exact block content after whitespace normalization, so paraphrased diagnostics receive no match. In Figure~\ref{fig:feedback-paraphrase}, the model identifies the same failing tactic and type conflict as Lean but phrases the error differently. Token F1 gives partial credit for their shared words and symbols. Overall block F1 ranges from 0.788 to 0.805 across prover families.

The model incorrectly accepts 260 of the 1{,}002 invalid-labelled programs, all without a predicted error block. Appendix~\ref{app:audit} gives the acceptance breakdown, including output-limit rejections. Invalid proofs also lead the model to generate excess feedback. It emits 1.53$\times$ as many blocks as Lean's trace on invalid programs, compared with 1.01$\times$ on valid ones. This is consistent with the model continuing after an error that ends the reference trace, as illustrated in Figure~\ref{fig:feedback-invalid}. The model supplies useful approximate correctness signals at the verdict and feedback levels, but exact failure diagnosis is less reliable. On invalid programs, error-presence agreement is 70.7\%; this measures agreement on whether an error is present, not reproduction of its exact message. Missed errors and excess feedback can complicate the use of predicted traces for proof repair.

\subsection{Pipeline throughput}
\paragraph{Speculative decoding.}
Speculative decoding accelerates feedback generation because much of the output repeats the input program. We serve the model with vLLM \citep{kwon2023vllm}, draft tokens by copying matching spans from the input, and let the model check those drafts \citep{vllmngram,leviathan2023speculative,chen2023speculative}. In a paired serving experiment with greedy decoding on 638 programs, drafting eight tokens at a time raises throughput from 1.04 to 3.17 programs/s, a 3.04$\times$ speedup. Accuracy changes from 0.851 to 0.850, with 99.2\% verdict agreement. Speculation also makes multiple-trace sampling faster. Separate serving runs give five-sample self-consistency throughput of 0.78 programs/s with speculation and 0.27 without it. On 638 programs evaluated under both configurations, accuracy is 0.886 in each and their verdicts agree on 94.7\% of programs.

\paragraph{FP8 quantization of the feedback model.}
Reducing numerical precision improves throughput while retaining similar prediction quality. We quantize the feedback model from 16-bit floating point (BF16) to 8-bit floating point (FP8), without a calibration dataset \citep{micikevicius2022fp8,vllmfp8}. With greedy decoding, accuracy is 0.852 with FP8 and 0.851 with BF16; verdicts agree on 98.3\% of programs. All reported trace-quality metrics differ by at most 0.005 on 989 matched programs. With speculation, FP8 raises throughput from 3.27 to 3.50 programs/s. The FP8 cascade reaches 0.877 accuracy at 0.279\,s per candidate.

\begin{table}[!htbp]
\caption{Neural throughput on one H100 and Lean throughput on CPUs. Feedback uses greedy decoding. Cascade timings include loading and model transitions; standalone timings exclude warmup and tokenization. The Lean REPL pool runs the benchmark verification pipeline with 64 workers and environment reuse.}
\label{tab:cost}
\centering\small
\begin{tabularx}{\linewidth}{@{}Yrr@{}}
\toprule
Evaluator & Candidates/s & Seconds/candidate \\
\midrule
ModernBERT & 264 & 0.0038 \\
One-token 8B & 4.9 & 0.204 \\
Feedback 4B, no speculation & 0.73 & 1.37 \\
Feedback 4B $+$ speculation & 3.27 & 0.306 \\
Cascade & 3.48 & 0.287 \\
Cascade, FP8 feedback stage & 3.59 & 0.279 \\
\midrule
Lean REPL pool, 64 workers & 0.835 & 1.20 \\
\bottomrule
\end{tabularx}
\par\smallskip\parbox{\linewidth}{\footnotesize Seconds per candidate is total batch time divided by batch size, not individual-request latency. Standalone rows use separate serving runs; the text reports the paired speculative-decoding comparison.}
\end{table}

\paragraph{End-to-end cascade throughput.}
With speculative decoding, the cascade processes 3.48 candidates per second in the recorded serving run (Table~\ref{tab:cost}). This includes loading the gate, scoring every candidate, switching models, and generating traces for the 64.8\% that pass the gate. The FP8 variant reaches 3.59 candidates per second. Appendix~\ref{app:deployment} gives the timing breakdown.

\paragraph{Comparison with Lean.}
How does this compare with running Lean itself? Our reference is the Lean pool that collected the benchmark: 64 parallel workers with environment reuse and incremental caching (Section~\ref{sec:benchmark}). It handled 78{,}074 verification requests in 26.0\,h, or 0.835 per second, so the cascade, at 3.48 candidates per second on one H100, is 4.2$\times$ faster. This compares whole pipelines: Lean returns exact verdicts, the cascade returns predictions, and the two schedule their work differently. Section~\ref{sec:grpo} measures what this speedup is worth in GRPO training.

\section{GRPO with hybrid rewards}
\label{sec:grpo}
\subsection{Reward scoring and training time}
After rollout generation, the hybrid protocol unloads the prover and uses the same eight H100 GPUs to score its 1{,}024 candidate proofs, with 128 candidates per GPU. Each GPU loads the one-token classifier, scores its candidates, then loads the feedback model for those that pass the gate. A 64-worker Lean pool checks candidates concurrently, with a 60\,s per-proof timeout. When all GPU predictions are ready, after 90\,s on average, we retain completed Lean verdicts and use cascade predictions for unfinished checks (Figure~\ref{fig:grpo-hybrid}). Deterministic precheck failures remain negative. The 90\,s includes model loading and unloading, warmup, inference, and synchronization. This cutoff is set by GPU evaluation time, not tuned; about 43\% of candidates then have Lean verdicts.

With Lean-only rewards and the same 60\,s per-proof timeout, scoring a batch takes about 420\,s. Reducing the per-proof timeout to 13\,s lowers batch scoring time to about 90\,s, matching the hybrid protocol. Generation and policy updates take approximately 630\,s in all three conditions, giving total step times of about 1{,}050\,s for standard Lean-only training and 720\,s for hybrid and short-timeout Lean-only training. Hybrid rewards thus reduce step time by about 31\%. These per-proof timeouts apply during GRPO training. Benchmark collection and final prover evaluation instead use a 180\,s timeout (Section~\ref{sec:benchmark}; Appendix~\ref{app:grpo}).

\subsection{Controlled GRPO training comparison}
We train an 8B prover for one epoch under three reward conditions: Lean-only with a 60\,s per-proof timeout, hybrid rewards, and Lean-only with a 13\,s per-proof timeout. All runs start from the same supervised checkpoint and use the same training data and GRPO settings, with eight rollouts for each of 128 problems per step on eight H100 GPUs. The 13\,s condition tests whether shortening Lean checks alone can match hybrid training at the same batch scoring time.

We evaluate the shared initial checkpoint and all three final checkpoints on all 488 miniF2F problems. Table~\ref{tab:grpo-training} reports means and standard deviations across five evaluations per fixed checkpoint for pass@1 and three for pass@32. Final proofs are checked with Lean independently of the learned reward predictions; Appendix~\ref{app:grpo} gives the evaluation protocol.

\begin{table}[htbp]
\caption{\textbf{Prover performance and training time.} Pass@1 and pass@32 on all 488 miniF2F validation and test problems, before and after one epoch of GRPO. Values are mean $\pm$ standard deviation across five evaluations for pass@1 and three for pass@32. Step times are approximate and include generation, reward scoring, and policy updates. Timeouts in the row labels apply to each proof during reward scoring.}
\label{tab:grpo-training}
\centering\small
\begin{tabularx}{\linewidth}{@{}Yrrr@{}}
\toprule
Checkpoint / reward source & Pass@1 (\%) & Pass@32 (\%) & Step time (s) \\
\midrule
Initial checkpoint (shared) & $53.8 \pm 0.6$ & $72.8 \pm 0.3$ & \NA \\
GRPO, Lean-only, 60\,s timeout & $62.4 \pm 0.6$ & $71.3 \pm 0.8$ & 1{,}050 \\
GRPO, hybrid rewards & $62.2 \pm 0.8$ & $72.4 \pm 0.4$ & 720 \\
GRPO, Lean-only, 13\,s timeout & $59.8 \pm 0.6$ & $69.7 \pm 1.1$ & 720 \\
\bottomrule
\end{tabularx}
\end{table}

All three reward conditions improve pass@1. Hybrid and standard Lean-only training finish at nearly the same level, 62.2\% and 62.4\%, while the 13\,s Lean-only condition reaches 59.8\%. At the same batch scoring time, hybrid rewards therefore give higher final pass@1 than shortening the Lean timeout alone.

Hybrid training also largely preserves the initial prover's pass@32: $72.4\pm0.4$\% after GRPO, compared with $72.8\pm0.3$\% before it. The small change is comparable to the variation across repeated evaluations. Lean-only training instead falls to $71.3\pm0.8$\% with the 60\,s timeout and $69.7\pm1.1$\% with the 13\,s timeout. Hybrid training finishes 1.1 and 2.7 percentage points above these conditions, respectively. The Lean-only pattern of rising pass@1 and falling pass@32 is consistent with distribution sharpening under GRPO, also observed in formal theorem proving by \citet{he2025rewarding}; the hybrid condition remains close to its initial pass@32 while gaining 8.4 points in pass@1.

\paragraph{Second seed.}
To check that this result is not a fluke of one run, we repeat hybrid training with a second seed. It reaches $62.7\pm0.7$\% pass@1 and $72.1\pm0.6$\% pass@32, close to the first run. Appendix~\ref{app:grpo-additional} reports this run together with ablations over reward sources.

\paragraph{Evaluator accuracy during training.}
Since the policy trains against the evaluator, it may learn to exploit the evaluator's errors. We track this on the rollouts of each hybrid run (Appendix~\ref{app:grpo-additional}). The cascade's accuracy drops early, from 87.9\% at step 25 to 81.9\% at step 125, as its false-positive rate rises from 17.8\% to 25.7\%. It then levels off: between steps 125 and 250, accuracy is nearly unchanged, ending at 81.7\%, while the false-positive rate rises by a further 3.2 points, to 28.9\%. Lean still supplies the reward for about 43\% of rollouts at all three steps. Weaker evaluators degrade much further, consistent with reward hacking: by step 250, the false-positive rate reaches 74.1\% for the one-token classifier and 99.6\% for ModernBERT.

\paragraph{Recovering timed-out rewards.}
Rechecking the 870 timeouts with a 600\,s budget resolves 797 cases, of which 463 (58.1\%) are valid; 73 remain unresolved (Appendix~\ref{app:grpo}). The cascade accepts 80.1\% of these valid proofs, so learned predictions can recover rewards that Lean withholds.

\section{Conclusion}
Learned Lean evaluators provide fast approximate verdicts and feedback for candidate proofs. Our cascade reaches 87.8\% accuracy at 3.5 candidates per second on one H100. Combining its predictions with completed Lean checks reduces GRPO step time by approximately 31\%, with nearly unchanged final miniF2F pass@1 and better preservation of the initial prover's pass@32 than Lean-only training. At the same scoring time, shortening the Lean timeout alone gives lower performance on both metrics. The \href{https://huggingface.co/collections/EPFL-LARA/learning-lean-feedback}{datasets and model checkpoints} and the \href{https://github.com/epfl-lara/learning-lean-feedback}{project repository} are publicly available.

\begin{ack}
This research is supported in part by the Swiss National Science Foundation grant number 10.006.623, the AI For Math Fund, administered by Renaissance Philanthropy and funded by XTX Markets, and the EPFL School of Computer and Communication Sciences. The authors declare no competing interests.
\end{ack}

{\small
\bibliographystyle{plainnat}
\bibliography{references}
}

\clearpage
\appendix
\section{Comparison of the Kimina Lean server with our Lean pipeline}
\label{app:kimina_comparison}
We compare our verification pipeline with Kimina Lean Server \citep{dossantos2025kimina} on a random subset of 10{,}000 proofs from NuminaMath-LEAN \citep{numina2024}. This provides an external check on the efficiency of our Lean baseline. Both systems use Lean v4.15.0, 64 Lean workers on a machine with 72 AMD EPYC 7R13 CPUs, and a 60\,s per-proof timeout, with a second attempt after a timeout. Kimina uses environment reuse and 10 HTTP workers, each submitting up to eight proofs per request. Our pipeline uses batches of 256 proofs.

Mean elapsed time per proof is 0.179\,s for our pipeline and 0.164\,s for Kimina, an 8.4\% reduction. Both compile the dataset's proof (the \texttt{answer} field). Our pipeline also elaborates the original theorem (the \texttt{question} field), checks that the proof preserves its statement, and rejects forbidden constructs such as \texttt{sorry}. This separate comparison uses a different workload and hardware from the public-prover collection measurements in Section~\ref{sec:results}.

\paragraph{CPU allocation.}
We run GRPO training on EPFL's shared RCP GPU cluster.\footnote{\url{https://www.epfl.ch/research/facilities/rcp/rcp-cluster-service-description/}} The 64-worker Lean pool uses the largest CPU allocation readily available to us. Larger systems use far more CPU cores: Kimina-Prover Preview reports 640 for verification during reinforcement learning \citep{wang2025kimina}, and DeepSeek-Prover-V1.5 reports thousands for Lean execution in its parallel tree search \citep{xin2024deepseekprover15}. Our pool already reuses loaded environments and caches compilation, and in a separate test, adding workers beyond 64 barely helps: going from 32 to 64 workers raises throughput by 15\%, but going to 96 adds less than 1\%, consistent with the slowest checks setting the batch time. Systems with much larger CPU budgets, like those above, may gain less from learned evaluation when verification is no longer a bottleneck; under CPU constraints like those of our shared university cluster, learned evaluation gives a clear speedup (Sections~\ref{sec:results} and~\ref{sec:grpo}).

\section{Reproducibility details}
\label{app:audit}
\paragraph{Provenance.}
The public-prover evaluations use saved predictions and timings from H100 GPUs, with each standalone evaluator and cascade pipeline using one GPU. GRPO scoring measurements use 128 candidates per H100; the training comparison uses eight H100s (Appendix~\ref{app:grpo}). The Lean collection reference aggregates 78{,}074 requests and 26.0\,h of verification wall time across the collection runs.

\paragraph{In-house candidate sampling.}
To generate additional proof candidates for evaluator training and the in-house validation set, we sample eight attempts per problem from the in-house 8B prover, using temperature 1.0, top-$p$ 0.95, and a limit of 32{,}768 generated tokens per attempt.

\paragraph{Benchmark reconstruction.}
Table~\ref{tab:runs} gives the six primary runs after exclusions and before deduplication. Removing 2{,}807 duplicates yields 68{,}932 programs, of which 40.8\% are valid. Verification wall times are 6.9/4.7\,h for Pythagoras on miniF2F/ProofNet, 4.5/1.7\,h for Goedel, and 4.8/3.2\,h for DeepSeek. Diagnostics are joined to generated candidates by index because ProofNet reuses some problem names.

Parameter selection uses the independent in-house validation data described in Section~\ref{sec:benchmark}.

\begin{table}[htbp]
\caption{Per-run composition before cross-run deduplication.}
\label{tab:runs}
\centering\small
\begin{tabularx}{\linewidth}{@{}Ylrrr@{}}
\toprule
Prover & Benchmark & Lean & Rows & Valid \\
\midrule
Pythagoras-Prover-4B & miniF2F & v4.15.0 & 14{,}903 & 0.574 \\
Pythagoras-Prover-4B & ProofNet & v4.15.0 & 10{,}335 & 0.039 \\
Goedel-Prover-V2-8B & miniF2F & v4.15.0 & 13{,}706 & 0.715 \\
Goedel-Prover-V2-8B & ProofNet & v4.15.0 & 6{,}205 & 0.155 \\
DeepSeek-Prover-V2-7B & miniF2F & v4.15.0 & 15{,}408 & 0.620 \\
DeepSeek-Prover-V2-7B & ProofNet & v4.15.0 & 11{,}182 & 0.124 \\
\bottomrule
\end{tabularx}
\end{table}

\paragraph{Lean version.}
Except for the version comparison below, all verification and annotation use Lean v4.15.0, released on January 4, 2025, with Mathlib. NuminaMath-LEAN is compiled and verified against Mathlib v4.15.0 \citep{numina2024}, while Goedel-Prover-V2 follows DeepSeek-Prover-V1.5 in using Lean 4.9 \citep{lin2025goedelv2}. This choice matches NuminaMath-LEAN and provides a common verification environment for our comparisons across provers. The 4.84M-program training corpus, its feedback annotations, and the evaluation benchmark are all produced with this version; a newer release would require re-verifying and re-annotating all of them. To check how much verdicts depend on the version, we verified 23{,}092 Pythagoras-Prover-4B candidates under both Lean v4.9.0-rc1 and v4.15.0: 98.9\% of verdicts agree, and 247 of the 259 changes turn a valid verdict invalid under v4.15.0. Changing versions can change verdicts and feedback, but the task remains predicting them for a program under a specified toolchain.

\paragraph{Verifier acceptance policy.}
Pre-checks exclude candidates that violate the collection policy before Lean runs. A candidate is rejected if it exceeds 262{,}144 characters, declares no theorem, or contains a forbidden construct after comments and strings are masked. Forbidden constructs are \texttt{sorry}, \texttt{admit}, \texttt{axiom}, \texttt{constant}, \texttt{opaque}, and \texttt{unsafe}. The import block must match the benchmark's required header, with no imports elsewhere in the program. Surviving candidates are compiled with a budget of 800{,}000 Lean heartbeats; proofs with unresolved placeholders are rejected.

Acceptance also requires a compiled declaration matching the expected theorem statement. The collection compares pretty-printed signatures and rejects mismatches even if the program otherwise compiles. This policy can conservatively reject a binder renaming that changes the printed signature. The pre-check does not block \texttt{native\_decide} or user metaprogramming such as \texttt{macro\_rules} and \texttt{elab}, so acceptance inherits the verifier's trust assumptions for those constructs. Timeouts and infrastructure failures produce no label. Evaluators receive the reconstructed candidate program, without a separate expected-statement input or Lean-version identifier. Matching the external target statement is enforced by the Lean collection harness.

\paragraph{Feedback-quality protocol.}
The main feedback sample is drawn from the evaluation benchmark. It contains 334 programs for each of the six prover--label combinations, giving 2{,}004 programs. Programs are sampled from saved greedy predictions using random seed 42. Reference annotation executes the programs through LeanInteract at Lean v4.15.0 with Mathlib, using the construction rules in Appendix~\ref{app:feedback-construction}.

Feedback metrics compare each predicted trace with the corresponding Lean annotation using the formulas in Appendix~\ref{app:feedback-metrics}. The scores in Table~\ref{tab:feedback} average per-program metrics over all 2{,}004 annotated programs, including truncated predictions; the valid and invalid columns use the stored benchmark labels to select their respective 1{,}002-program subsets.

The acceptance breakdown applies the evaluator's output-limit policy as well as the trace parser. Of 1{,}002 invalid-labelled programs, 714 contain a predicted error and are rejected. Of the remaining 288, 28 hit the output limit and 260 are accepted. Fresh Lean executions during annotation disagree with four stored labels; runs without a returned verdict are excluded from this comparison.

\paragraph{Parser and generation limits.}
For untagged traces, we predict invalid when the trace contains an error marker, ends in a terminal proof-state block, or reaches the output-token limit. Error detection uses the case-insensitive pattern in Appendix~\ref{app:feedback-metrics}. A terminal state is a complete feedback block containing \texttt{-- proof state:} with no non-whitespace source after it, after removing the end-of-text suffix. An explicit validity tag, if generated, overrides the annotation-derived verdict; when several appear, the last is used. Output-limit stops still force rejection. Such tags are absent from the training targets. The feedback evaluator rejects overlength inputs; the classifiers instead truncate them. Across the reported generation settings, 1.5--2.0\% of traces reach the output-token limit and are scored invalid.

\paragraph{Continued pretraining.}
The approximately 130M-token continued-pretraining corpus comprises Lean source and documentation from Mathlib and Lean Pool\footnote{\url{https://github.com/Vilin97/lean-pool}}. All three models train for one epoch with AdamW, cosine learning-rate decay, 3\% warmup, BF16 precision, and gradient clipping at norm 1.0. The encoder uses masked-token prediction with 30\% masking, an 8{,}192-token context, effective batch size 32, and learning rate $3\times10^{-5}$. Both Qwen models use next-token prediction on packed 16{,}384-token sequences, effective batch size eight, and learning rate $5\times10^{-6}$.

\paragraph{Supervised evaluator training.}
The three models draw from the same compiler-labelled program pool, with model-specific filtering. The one-token dataset contains 4{,}788{,}241 examples packed into 130{,}000 sequences; the feedback dataset contains 4{,}836{,}380 examples packed into 443{,}760 sequences. At an effective batch size of eight, these correspond to 16{,}250 and 55{,}470 optimizer steps. The encoder processes approximately 4.82M examples over 75{,}251 steps. All supervised training uses BF16 precision. Both Qwen models use a cosine learning-rate schedule with 5\% warmup and pack multiple examples into 32{,}768-token sequences. The one-token model retains the beginning and end of inputs that exceed its context limit. Both Qwen models receive raw program text without a chat template at training and inference. For the feedback model, the loss is applied only to the annotated response, not to the input prompt.

\paragraph{Evaluator-training compute.}
Including continued pretraining and supervised training, the encoder takes approximately 12\,h on one H100, the one-token 8B model 24\,h on eight H100s, and the feedback 4B model 67\,h on eight H100s. These runs use approximately 12, 192, and 536 H100 GPU-hours, respectively, for a combined 740 H100 GPU-hours. This total covers the three evaluator-training runs; it excludes data collection, inference, GRPO training, and preliminary experiments.

\paragraph{Timing boundaries.}
Neural hardware is one NVIDIA HGX H100-SXM5-80GB GPU, serving with vLLM 0.12.0, PyTorch 2.9.0+cu129, and Transformers 4.57.3 (the gate runs as a Transformers forward pass with FlashAttention-2). The compiler reference uses dual-socket AMD EPYC 7543 nodes (64 cores, 128 threads, 1{,}008\,GiB RAM) with 64 concurrent Lean verification jobs, reuse of loaded environments, incremental compilation caching, and a 180\,s per-command timeout; the machine type is the same across collection runs.

Pipeline timing includes the initial model load, gate scoring, model transition, feedback generation, and driver overhead. The transition is the gate unload plus feedback-model load. These startup timers cover different work: the gate loads a Transformers model, while the feedback timer covers vLLM engine initialization, including its setup and warmup. The current pipeline takes 979.7\,s, of which 8.9\,s is driver overhead beyond the timed model stages. Table~\ref{tab:cost} reports amortized batch throughput. The Lean reference measures the 64-worker REPL verification pipeline used to collect the benchmark, including its acceptance checks and environment reuse; its throughput is 0.835 candidates/s, or 1.20\,s per candidate averaged over the collection workload.

\paragraph{Source preservation.}
Source-position matching uses the model's emitted program text, so we also check whether that text preserves the input. In the recorded source-preservation audit, 54.9\% of generations reproduce the full input, 44.7\% reproduce an exact prefix, and 0.41\% contain other changes. The prefix category compares emitted source with the corresponding input prefix, allowing traces that stop at a predicted error. The audit measures source preservation separately from feedback quality.

\paragraph{Complete feedback comparisons.}
Figures~\ref{fig:feedback-paraphrase} and~\ref{fig:feedback-invalid} compare saved model predictions with annotations obtained from Lean v4.15.0 on two invalid proofs. In the first, the model predicts an error at the same tactic as Lean and states the same type conflict, but its message differs. The first difference is the message kind: Lean reports an \emph{application type mismatch}, because the anonymous constructor elaborates to an application of \texttt{Exists.intro} whose argument has the wrong type, while the model predicts a plain \emph{type mismatch} and omits the application. Exact block matching counts this error block as a miss even though both goal-state blocks match, so token F1 exceeds block F1 for this program. In the second, the model misses the error and continues annotating. Each figure displays both traces in full and reports the feedback metrics for that individual program; aggregate quality appears in Table~\ref{tab:feedback}.

\section{Construction of feedback traces}
\label{app:feedback-construction}
Feedback traces serialize Lean's tactic states and diagnostics as compact comments in the source program. The same format defines the feedback model's prediction target and the reference annotations used for feedback-quality evaluation.

\paragraph{Source preparation and trace collection.}
For fresh annotation, the batch pipeline removes existing feedback and proof-state blocks, then places inline declaration proofs of the form \texttt{:= by tactic} onto an indented tactic line. This gives the first tactic its own insertion point. It executes the resulting program through LeanInteract with tactic tracing enabled. Tactic records provide the goal-state text and source span; compiler messages provide severity, message text, and source positions. If multiple tactic records start on the same source line, only the first record returned for that line is retained. Messages are grouped by the ending line of their reported source span. The trace is therefore aligned to source lines rather than containing a separate block for every nested tactic invocation.

\paragraph{Placement and block merging.}
The annotator scans source lines in order. Before a retained tactic's starting line, it inserts a comment beginning with \texttt{/- <feedback>} and \texttt{-- proof state:}. It then copies the source line and appends any diagnostic feedback assigned to that line. Comment indentation follows the associated source line, and each block ends with \texttt{</feedback> -/}. When a proof-state payload immediately follows an existing feedback block, the formatter removes the preceding closing delimiter and appends the new payload inside that block. This avoids adjacent wrappers, for example when a diagnostic is followed immediately by the next tactic's state. The metrics in Appendix~\ref{app:feedback-metrics} score these serialized blocks.

\paragraph{Hypothesis deduplication.}
The formatter separates each printed proof state into hypothesis declarations and goal/case lines. It joins continuation lines within a declaration and expands grouped names, such as \texttt{x y : T}, into individual name--value entries. A dictionary, initially empty for each annotated program, stores the last printed value for every hypothesis name. A newly encountered declaration is printed as \texttt{h : P}; an unchanged declaration is omitted. If the same name later has a different printed value, the formatter emits \texttt{h : Q (was P)} and updates the dictionary. Goal targets, case labels, and other retained non-declaration lines are emitted even when they repeat.

The dictionary persists throughout the source scan and compares printed names and values; it does not perform semantic equivalence checking or track Lean declaration identities. Names absent from a later state are not removed from the dictionary, and no explicit deletion marker is emitted. Thus, a compact state comment records new or changed hypotheses together with the displayed goals; it is not a standalone reproduction of the full local context. Deduplication applies to parsed proof-state declarations, not to hypothesis text embedded inside compiler error messages.

\paragraph{Diagnostics and stopping.}
Each retained diagnostic is serialized as \texttt{-- type: severity, msg: text}. The formatter skips warnings by default and removes message lines beginning with \texttt{note: this linter can be disabled with}; other diagnostic text is retained, including multiline messages. The batch renderer emits no diagnostic block if filtering leaves no message text. After emitting the first error-bearing diagnostic block, it ends the annotated source prefix, omitting later source lines and feedback. This stopping rule belongs to the annotation pipeline. For a completed valid proof, no synthetic final \texttt{no goals} message or separate validity tag is appended; terminal goal-state text is not used as an extra success marker.

\section{Feedback-quality metrics}
\label{app:feedback-metrics}
The four metrics in Table~\ref{tab:feedback} compare the model's predicted annotated program $\widehat z_i$ with Lean's reference annotation $z_i$ for the same input program $i$.

\paragraph{Block extraction and normalization.}
From each annotated program, we extract complete blocks delimited by \texttt{/- <feedback>} and \texttt{</feedback> -/}, allowing whitespace around the tags and matching the tags case-insensitively. Unclosed blocks are not extracted. We remove the delimiters and normalize each block's content by stripping leading and trailing whitespace and replacing every internal whitespace run with one space. Content matching remains case-sensitive and retains all other characters, including comment markers.

Let $B_i$ and $\widehat B_i$ be the multisets of normalized reference and predicted block contents. For each block, its anchor $a$ is the number of non-empty lines preceding it in its own annotated program, after removing earlier feedback blocks. Let $A_i$ and $\widehat A_i$ be the corresponding multisets of pairs $(a,b)$, where $b$ is the normalized block content. Thus, anchors are computed separately from the reference and predicted source text.

\paragraph{Multiset precision, recall, and F1.}
For a reference multiset $G$ and a predicted multiset $P$, write $c_G(u)$ and $c_P(u)$ for the multiplicity of item $u$, and let $|G|$ and $|P|$ count items with multiplicity. Their overlap is
\begin{equation}
O(G,P)=\sum_{u\in\operatorname{supp}(G)\cup\operatorname{supp}(P)}
\min\{c_G(u),c_P(u)\}.
\label{eq:feedback-overlap}
\end{equation}
For non-empty denominators, precision is $\operatorname{Prec}(G,P)=O(G,P)/|P|$ and recall is $\operatorname{Rec}(G,P)=O(G,P)/|G|$. If both multisets are empty, precision and recall are both 1; if exactly one is empty, both are 0. The harmonic mean is equivalently
\begin{equation}
F(G,P)=
\begin{cases}
1, & |G|+|P|=0,\\
\displaystyle\frac{2O(G,P)}{|G|+|P|}, & |G|+|P|>0.
\end{cases}
\label{eq:feedback-f1}
\end{equation}
Repeated blocks or tokens can therefore match only as many times as they occur on both sides.

\paragraph{Block and token scores.}
Let $T_i$ and $\widehat T_i$ be the multisets of tokens pooled over all normalized block contents on each side. Tokenization uses the Unicode-aware Python regular expression \verb!\w+|[^\w\s]!: each token is either a maximal run of word characters (Unicode alphanumeric characters or underscores), or one non-word, non-whitespace character. Tokens are case-sensitive; their order and block boundaries are discarded. The three F1 metrics are
\begin{align}
\operatorname{BlockF1}_i &= F(B_i,\widehat B_i),\label{eq:feedback-block-f1}\\
\operatorname{AnchoredBlockF1}_i &= F(A_i,\widehat A_i),\label{eq:feedback-anchored-f1}\\
\operatorname{TokenF1}_i &= F(T_i,\widehat T_i).\label{eq:feedback-token-f1}
\end{align}
Block F1 ignores block order and placement; anchored block F1 additionally requires the same anchor. Token F1 gives partial credit for shared tokens even when an entire diagnostic does not match. The empty-multiset convention applies separately to each metric.

\paragraph{Error-presence agreement.}
Define $E(z)$ to be 1 if the case-insensitive regular expression \verb!--\s*type\s*:\s*error\b! matches anywhere in $z$, and 0 otherwise. The implementation searches the full annotated text, rather than only the extracted complete blocks. Per-program agreement is
\begin{equation}
\operatorname{ErrorAgreement}_i
=\mathbf{1}\{E(z_i)=E(\widehat z_i)\}.
\label{eq:feedback-error-agreement}
\end{equation}
Both predicting an error when one is present and predicting no error when none is present count as agreement. This score compares error markers independently of the stored validity label and the output-limit policy used for verdict prediction.

\paragraph{Averaging.}
For any of the four per-program metrics $m_i$ and a scored subset $S$, the reported value is
\begin{equation}
\overline m_S=\frac{1}{|S|}\sum_{i\in S}m_i.
\label{eq:feedback-average}
\end{equation}
Every program has equal weight: F1 is averaged after per-program computation, rather than computed from pooled corpus-wide counts or averaged precision and recall. The scorer excludes rows whose reference annotation is missing or blank; a non-empty annotated program containing no feedback blocks remains eligible. Table~\ref{tab:feedback} includes all 2{,}004 sampled programs, with no exclusions for missing reference annotations. The valid and invalid subsets each contain 1{,}002 programs; truncated predictions remain included and are scored using their emitted text.

\section{GRPO evaluation protocol}
\label{app:grpo}
\paragraph{Cohort and reward labels.}
The replay cohort contains 24{,}576 rollouts, partitioned into 23{,}489 with an original Lean verdict (63.9\% valid), 870 timeouts, and 217 without extractable Lean code. Verdict accuracy uses only the 23{,}489 labelled candidates; timeout rechecks form a separate audit. On these candidates, feedback-model accuracy is 85.3\% and greedy-cascade accuracy is 86.2\%. The feedback model retains 89.9\% of valid proofs and accepts 22.8\% of invalid ones; the cascade's rates are 85.1\% and 11.6\%. The feedback model uses one greedily decoded trace with speculative acceleration. The cascade uses the one-token threshold of 0.5 followed by the same feedback rule. The hybrid protocol uses cascade predictions for unfinished checks: the one-token classifier screens each candidate, and only its positive predictions proceed to the 4B feedback model. Both stages must accept a candidate for the cascade to predict validity.

In this cohort, $870/24{,}576=3.54\%$ of rollouts time out and $217/24{,}576=0.88\%$ contain no extractable code, so $1{,}087/24{,}576=4.42\%$ lack an original verdict. Figure~\ref{fig:grpo-hybrid} excludes missing-code cases and reports $870/24{,}359=3.57\%$ as 3.6\% timeouts. These denominators distinguish timeouts from other missing rewards.

\paragraph{Hybrid replay and timing.}
For candidate $i$, let $d_i$ indicate a deterministic precheck failure, $y_i$ be its Lean verdict, $c_i$ the time at which that verdict becomes available relative to the start of the reward window, and $\widehat y_i$ the cascade's prediction. Let $T_{\mathrm{eval}}$ be the time when all LLM predictions are ready. The hybrid verdict is
\begin{equation}
\widetilde y_i(T_{\mathrm{eval}})=
\begin{cases}
0, & \text{if }d_i=1,\\
y_i, & \text{if }d_i=0\text{, a Lean verdict is available, and }c_i\le T_{\mathrm{eval}},\\
\widehat y_i, & \text{otherwise.}
\end{cases}
\end{equation}
Missing code, forbidden constructs, and other deterministic precheck failures cannot receive a positive learned correctness reward. The replay uses saved Lean request-completion times, with the first reward request as the time origin. It sets $T_{\mathrm{eval}}=\hybridcutoff\,\mathrm{s}$, the mean LLM evaluation time, and preserves the recorded Lean completion schedule. This replay boundary assumes all evaluator predictions are ready then; Lean checks proceed concurrently throughout the window. Re-evaluation on the full cohort gives 92.5\% agreement on the 23{,}489 candidates with original reference verdicts; timeouts and missing-code cases are excluded from this denominator.

During training, the reward window ends when all evaluator shards have finished. Each GPU receives 128 of the 1{,}024 rollouts, and Lean verdicts returned by LLM completion take precedence over cascade predictions. The mean \hybridevalmean\,s evaluation time includes loading and unloading both models, setup and warmup, classifier and feedback inference, and synchronization across eight H100s. At completion, about 43\% of the candidates use Lean verdicts and the rest use cascade predictions (Appendix~\ref{app:grpo-additional}); unfinished Lean checks are cancelled. The reported step-time breakdown includes approximately 630\,s for candidate generation and policy updates (including backpropagation and optimizer updates), plus 420\,s for Lean-only reward scoring or \hybridevalmean\,s for hybrid scoring. Thus
\begin{equation}
T_{\mathrm{Lean}}\approx630+420=1{,}050\,\mathrm{s},\qquad
T_{\mathrm{hybrid}}\approx630+90=720\,\mathrm{s}.
\end{equation}
The reduction is approximately 330\,s per step, or $100(1{,}050-720)/1{,}050\approx31.4\%$, reported as 31\% in the main text. This reduction corresponds to the reported eight-H100 configuration. The final comparisons evaluate checkpoints trained for one epoch on the same data.

\paragraph{Extended-budget timeout audit.}
Rechecking all 870 timeouts with a 600\,s budget resolves 797 cases: 463 valid and 334 invalid. The resolved valid fraction is $463/797=58.1\%$. The cascade predicts valid for 371 of the 463 valid proofs, yielding $371/463=80.1\%$ recall. The 73 unresolved cases remain unlabelled. These recovered labels do not enter the original replay's accuracy denominator.

\paragraph{Matched GRPO runs.}
All three reward conditions start from the same in-house 8B prover, supervised on correct proofs from the collected datasets. They train for one epoch on approximately 21.8K filtered proof-completion problems from Goedel-Prover-V2 RL data\footnote{\url{https://huggingface.co/datasets/Goedel-LM/RL_dataset_V2}} \citep{lin2025goedelv2}. Following the Goedel-Prover-V2 filtering rule, we retain problems whose empirical per-problem pass rate $p$ satisfies $0<p\le0.75$. Each training batch contains 128 prompts and eight rollouts per prompt. All runs use the optimizer, sampling configuration, and other reward settings in Table~\ref{tab:grpo-config}. Standard Lean-only and hybrid scoring use a 60\,s per-proof Lean timeout; the short-timeout Lean-only control uses 13\,s. The resulting batch scoring times are approximately 420\,s, 90\,s, and 90\,s, respectively. All runs use eight H100 GPUs; the hybrid run reuses the training allocation for evaluator inference and preserves optimizer state across model transitions.

\begin{table}[htbp]
\caption{GRPO settings shared by all three reward conditions. Each run trains for one epoch; reward-scoring timeouts are specified above.}
\label{tab:grpo-config}
\centering\small
\begin{tabularx}{\linewidth}{@{}lY@{}}
\toprule
Setting & Value \\
\midrule
Training prompts / rollouts per prompt & 128 / 8 \\
PPO minibatch / actor microbatch & 128 prompts / 1 sequence per GPU \\
PPO clip ratio / update epochs per rollout batch & 0.2 / 1 \\
Optimizer / $(\beta_1,\beta_2)$ & AdamW / $(0.9,0.999)$ \\
Weight decay / gradient clipping & 0.01 / 1.0 \\
Learning rate / warmup & $3\times10^{-6}$ / 25 updates \\
Learning-rate schedule & Cosine decay to 10\% of the initial rate \\
KL regularization & Low-variance estimator; coefficient 0.01 in the loss \\
Entropy coefficient & 0 \\
Loss aggregation & Mean token loss per response, then mean over responses \\
Malformed output / improper termination penalties & 0.1 / 0.15 \\
Verification-timeout / output-filter penalties & 0.02 / 0.5 \\
Rollout temperature / top-$p$ & 1.0 / 0.95 \\
Rollout tensor parallelism & 1 \\
Prompt / output / total context limit & 2{,}048 / 30{,}720 / 32{,}768 tokens \\
Data seed & 1234 \\
Training allocation & Eight H100 GPUs \\
\bottomrule
\end{tabularx}
\end{table}

\paragraph{Estimated GRPO training compute.}
One epoch over approximately 21.8K problems with 128 prompts per step corresponds to about 170 training steps. Applying the step times above gives approximately 50\,h for standard Lean-only training and 34\,h each for hybrid and 13\,s Lean-only training on eight H100s, or approximately 400, 270, and 270 H100 GPU-hours. Together with the 740 H100 GPU-hours for evaluator training, the three evaluator-training runs and three GRPO runs require an estimated 1{,}680 H100 GPU-hours. These estimates use training exposure and step times; data collection, evaluation, and preliminary experiments are outside this total. The additional runs in Appendix~\ref{app:grpo-additional} and training beyond the one-epoch checkpoints are also excluded.

\paragraph{miniF2F evaluation.}
We evaluate the shared initial checkpoint and all three final checkpoints on all 488 miniF2F problems (244 validation and 244 test problems). For pass@1, we perform five evaluations per checkpoint, generating one candidate per problem in each repeat, for 2{,}440 candidates per checkpoint. For pass@32, we perform three evaluations per checkpoint, generating 32 candidates per problem in each repeat, for 46{,}848 candidates per checkpoint. Decoding uses temperature 0.3 for pass@1 and 0.7 for pass@32; both use top-$p$ 0.95 and a cap of 32{,}768 generated tokens per candidate. All checkpoints use the same Lean version, imports, and acceptance policy, with a 180\,s verification timeout per proof for both metrics. This benchmark-evaluation budget is separate from the 60\,s and 13\,s timeouts used for GRPO reward scoring. Final evaluation uses the Lean verification harness independently of learned reward predictions.

For checkpoint $a$, repeat $r$, and sample count $k\in\{1,32\}$, let $c_{ar}^{(k)}$ be the number of problems with at least one accepted proof among their $k$ generated candidates. The success rate in that repeat is
\begin{equation}
p_{ar}^{(k)}=\frac{c_{ar}^{(k)}}{488}.
\end{equation}
Unsolved problems remain in the denominator, including those with only rejected candidates, missing-code outputs, or timeouts. With $R_1=5$ and $R_{32}=3$ evaluations, the reported mean on the percentage scale is
\begin{equation}
\overline P_a^{(k)}=\frac{100}{R_k}\sum_{r=1}^{R_k}p_{ar}^{(k)}.
\end{equation}
We report this mean with the standard deviation across repeats, also on the percentage scale. The standard deviations quantify variation across repeated decoding and evaluation runs for each fixed checkpoint. Table~\ref{tab:grpo-training} reports all four checkpoints under both metrics.

\section{Additional GRPO runs}
\label{app:grpo-additional}
\paragraph{Second seed and reward sources.}
We repeat the hybrid run with a different random seed and train three variants that change only the reward source: hybrid rewards with the one-token classifier at its calibrated threshold of 0.86, hybrid rewards with ModernBERT at threshold 0.5, and cascade predictions for every candidate without Lean. All runs otherwise match the main hybrid run's initial checkpoint, training data, GRPO settings (Table~\ref{tab:grpo-config}), and miniF2F evaluation protocol (Appendix~\ref{app:grpo}); all reported miniF2F results use the one-epoch checkpoint (about 170 steps). In each hybrid run, Lean checks stop when that run's evaluator finishes. Each GRPO row in Table~\ref{tab:grpo-additional} represents one training run.

\begin{table}[!htb]
\caption{Additional GRPO runs on all 488 miniF2F problems after one epoch. Values are mean $\pm$ standard deviation across five evaluations for pass@1 and three for pass@32. The first two rows repeat Table~\ref{tab:grpo-training} for reference.}
\label{tab:grpo-additional}
\centering\small
\begin{tabularx}{\linewidth}{@{}Yrr@{}}
\toprule
Checkpoint / reward source & Pass@1 (\%) & Pass@32 (\%) \\
\midrule
Initial checkpoint (shared) & $53.8 \pm 0.6$ & $72.8 \pm 0.3$ \\
GRPO, hybrid rewards (cascade) & $62.2 \pm 0.8$ & $72.4 \pm 0.4$ \\
\midrule
GRPO, hybrid rewards (cascade), second seed & $62.7 \pm 0.7$ & $72.1 \pm 0.6$ \\
GRPO, hybrid rewards (one-token, threshold 0.86) & $57.6 \pm 0.8$ & $70.3 \pm 0.4$ \\
GRPO, hybrid rewards (ModernBERT, threshold 0.5) & $48.4 \pm 0.6$ & $63.1 \pm 0.4$ \\
GRPO, cascade predictions only, no Lean & $59.0 \pm 0.6$ & $70.7 \pm 0.1$ \\
\bottomrule
\end{tabularx}
\end{table}

The second seed gives results close to the first run. The one-token and ModernBERT variants finish below the cascade on both metrics, and ModernBERT falls below the initial checkpoint. Cascade predictions without Lean reach 59.0\% pass@1, 3.2 points below hybrid training with the same evaluator; because Lean runs concurrently with GPU evaluation, its completed verdicts are retained without waiting for further checks. These runs compare complete reward pipelines with evaluator-dependent cutoffs and different observed Lean shares (Table~\ref{tab:grpo-drift}); step times for the additional runs are not reported.

\paragraph{Evaluator accuracy during training.}
To track how evaluator accuracy changes as the policy trains, we continue the three hybrid runs past the one-epoch checkpoint of about 170 steps to step 250. At steps 25, 125, and 250, we compare each evaluator's verdicts on all rollouts of that step in its own run with Lean labels obtained afterwards, including rollouts that received Lean rewards during training (Table~\ref{tab:grpo-drift}). The false-positive rate (FPR) is the fraction of Lean-invalid rollouts that the evaluator accepts; the Lean share is the fraction of rollouts whose Lean verdict arrives before the evaluator finishes. Evaluator errors can affect rewards only where no Lean verdict is available at the cutoff and deterministic prechecks do not determine the reward; these all-rollout metrics do not isolate that subset.

\begin{table}[!htb]
\caption{Evaluator accuracy on the rollouts of hybrid GRPO training, against Lean labels. Each evaluator is measured in its own hybrid run. All values are percentages.}
\label{tab:grpo-drift}
\centering\small
\begin{tabularx}{\linewidth}{@{}Yrrrr@{}}
\toprule
Evaluator in hybrid run & Step & Accuracy & FPR & Lean share \\
\midrule
Cascade & 25 & 87.9 & 17.8 & 42.8 \\
 & 125 & 81.9 & 25.7 & 42.8 \\
 & 250 & 81.7 & 28.9 & 43.7 \\
\midrule
One-token 8B (threshold 0.86) & 25 & 81.1 & 37.0 & 41.9 \\
 & 125 & 68.0 & 56.9 & 40.2 \\
 & 250 & 59.6 & 74.1 & 40.1 \\
\midrule
ModernBERT (threshold 0.5) & 25 & 58.2 & 87.6 & 36.7 \\
 & 125 & 29.0 & 99.3 & 28.5 \\
 & 250 & 33.7 & 99.6 & 33.4 \\
\bottomrule
\end{tabularx}
\end{table}

All three evaluators lose accuracy as the policy changes, at very different rates. The cascade's accuracy falls from 87.9\% to 81.9\% by step 125 and remains close, at 81.7\% at step 250, while its FPR rises from 17.8\% to 28.9\%. The one-token classifier's FPR rises from 37.0\% to 74.1\%. ModernBERT accepts nearly all invalid rollouts at steps 125 and 250, consistent with the policy exploiting its errors; its one-epoch checkpoint also falls below the initial checkpoint (Table~\ref{tab:grpo-additional}).

\paragraph{Reward finality and evaluator updates.}
The evaluator stays fixed during GRPO, and rewards are not revised after the cutoff: unfinished Lean checks are cancelled, so the reported step times include no correction stage. The Lean verdicts collected during hybrid training could instead be used to update the evaluator between training cycles, which may counter the drift in Table~\ref{tab:grpo-drift}; we leave this to future work.

\section{Additional evaluation details}
\label{app:deployment}

The fraction of valid candidates affects how trustworthy a positive prediction is. Let $\pi$ denote that fraction, and let $\operatorname{TPR}(t)$ and $\operatorname{FPR}(t)$ be the true-positive and false-positive rates at threshold $t$. Precision, also called positive predictive value (PPV), is
\begin{equation}
\operatorname{PPV}(t;\pi)=
\frac{\pi\operatorname{TPR}(t)}{\pi\operatorname{TPR}(t)+(1-\pi)\operatorname{FPR}(t)}.
\end{equation}
Thresholds are selected on the in-house validation set using a grid with spacing 0.01, then evaluated on the evaluation benchmark. Maximizing validation MCC selects a one-token threshold of 0.86, giving test accuracy 0.831 and MCC 0.683. Matching the cascade's validation false-positive rate selects 0.89, giving 0.836 accuracy and 0.685 MCC. Matching its valid recall selects 0.66, giving 0.791 accuracy and 0.640 MCC. All three comparisons favor the default cascade, which reaches 0.878 accuracy and 0.772 MCC. Selecting the cascade gate threshold on validation data gives 0.82, with test accuracy 0.887 and MCC 0.778. This reduces valid recall from 0.974 to 0.937.

Bootstrap intervals account for dependence among candidates sharing a problem name. We group evaluated programs by benchmark and problem name and resample the groups 5{,}000 times with replacement using random seed 42. The 2.5th and 97.5th percentiles give a cascade-accuracy interval of 0.863--0.892. Using the same resamples for the default and self-consistency cascades gives an accuracy-gain interval of 0.004--0.017.

Pooled AUROC includes comparisons between candidates from different benchmarks. Since the valid fractions differ substantially between miniF2F and ProofNet, this mixture can affect the pooled ranking score.

\paragraph{Budgeted compilation replay.}
The replay measures how many candidate pools yield a valid proof at fixed compilation budgets. Candidates are grouped by benchmark, run, and problem name. Each group is ranked either by the one-token score or by cascade verdict with the one-token score breaking ties. We compare both with the exact expected coverage under random ordering. Mean pool sizes are 28.5 for miniF2F and 27.3 for ProofNet. A valid candidate exists in 1{,}125 of 1{,}461 miniF2F pools and 173 of 1{,}001 ProofNet pools; coverage is measured over those pools. ProofNet reuses 19 problem names, so some pools merge distinct theorems.

Cascade ranking improves coverage most at small budgets. On ProofNet, one candidate per pool uses 3.7\% of calls and recovers 57.2\% of pools containing a valid proof, compared with 52.6\% for one-token ranking and 47.9\% for random ordering. At two candidates, cascade and random coverage are 72.3\% and 63.2\%; at eight candidates they are 89.6\% and 87.7\%. On miniF2F, one candidate gives 83.9\% versus 79.3\% random coverage, and four give 95.2\% versus 92.2\%. These are compilation-call budgets; elapsed-time savings also depend on evaluator cost, proof length, and Lean scheduling.

\paragraph{Cascade accounting.}
Pipeline cost combines gate scoring, selective feedback generation, and model overhead. For $N$ programs and gate-positive fraction $q$, the average elapsed time per candidate is
\begin{equation}
\overline C_{\mathrm{cascade}}=\overline C_{\mathrm{gate}}
+q\,\overline C_{\mathrm{trace}\mid\mathrm{gate+}}
+\frac{C_{\mathrm{load/swap}}+C_{\mathrm{other}}}{N}.
\end{equation}
In the measured pipeline, $q=0.648$, gate scoring takes 420.2\,s, and feedback generation takes 521.8\,s for 2{,}212 candidates. Thus $\overline C_{\mathrm{gate}}=0.123$\,s per input and $\overline C_{\mathrm{trace}\mid\mathrm{gate+}}=0.236$\,s per gate-positive input. Initial gate loading takes 3.4\,s, the model transition 25.4\,s, and other driver work 8.9\,s. Together these sum to the recorded 979.7\,s, or 0.287\,s per candidate. Standalone model timings use their own batch configurations; the cascade total uses the stages measured together in this pipeline.

\section{AI use and reproducibility statements}
\label{app:statements}
\paragraph{AI use statement.}
  Generative AI tools were used to edit the manuscript and prepare the
  schematic hybrid-evaluation visualization. The figure was checked against
  the reported aggregate timing and coverage values; its intermediate
  profiles and phase durations are illustrative. The use of language models
  as components of the research methodology is described in
  Sections~\ref{sec:benchmark}--\ref{sec:grpo}. We reviewed all AI-assisted
  work and take responsibility for the final content of this paper.

\paragraph{Reproducibility statement.}
  Sections~\ref{sec:benchmark}--\ref{sec:grpo} describe the datasets,
  models, training procedures, and evaluation protocols. The appendices
  provide additional details on verification, feedback construction,
  metrics, GRPO configuration, and deployment. The datasets and model
  checkpoints are publicly available at the links given in Section~\ref{sec:intro}.

\begin{figure}[!ht]
\centering
\begin{minipage}[t]{0.49\textwidth}
\vspace{0pt}
\centering\small\textbf{Lean annotation}
\par\smallskip
\raggedright\fontsize{8}{9.2}\selectfont
\begin{Verbatim}[commandchars=\\\{\},breaklines,breakanywhere=false,breaksymbolleft={},breaksymbolright={}]
\LeanKeyword{import} Mathlib
\LeanKeyword{set_option} maxHeartbeats 800000
\LeanKeyword{open} Fintype Set Real Ideal Polynomial
\LeanKeyword{open} \LeanKeyword{scoped} BigOperators
\LeanKeyword{open} Fintype Set Real Ideal Polynomial
\LeanKeyword{open} \LeanKeyword{scoped} BigOperators
\LeanKeyword{theorem} exercise_4_3_1 \{R : \LeanKeyword{Type}*\}
    [CommRing R] (a : R) :
    ∃ I : Ideal R, \{x : R | x*a=0\} = I := \LeanKeyword{by}
  \color{leanDelimiter}/- <feedback>
  \color{leanGoal}-- proof state:
  \color{leanGoal}R : Type u_1
  \color{leanGoal}inst✝ : CommRing R
  \color{leanGoal}a : R
  \color{leanGoal}⊢ ∃ I, \{x | x * a = 0\} = ↑I
  \color{leanDelimiter}</feedback> -/
  \LeanKeyword{have} h_main : ∃ (I : Ideal R),
      \{x : R | x*a=0\} = I := \LeanKeyword{by}
    \color{leanDelimiter}/- <feedback>
    \color{leanGoal}-- proof state:
    \color{leanGoal}⊢ ∃ I, \{x | x * a = 0\} = ↑I
    \color{leanDelimiter}</feedback> -/
    \LeanKeyword{refine'} ⟨\{x : R | x * a = 0\}, _⟩
    \color{leanDelimiter}/- <feedback>
    \color{leanError}-- type: error, msg:
      \color{leanError}application type mismatch
      \color{leanError}Exists.intro \{x | x * a = 0\}
    \color{leanError}argument
      \color{leanError}\{x | x * a = 0\}
    \color{leanError}has type
      \color{leanError}Set R : Type u_1
    \color{leanError}but is expected to have type
      \color{leanError}Ideal R : Type u_1
    \color{leanDelimiter}</feedback> -/
\end{Verbatim}
\end{minipage}\hfill
\begin{minipage}[t]{0.49\textwidth}
\vspace{0pt}
\centering\small\textbf{Model prediction}
\par\smallskip
\raggedright\fontsize{8}{9.2}\selectfont
\begin{Verbatim}[commandchars=\\\{\},breaklines,breakanywhere=false,breaksymbolleft={},breaksymbolright={}]
\LeanKeyword{import} Mathlib
\LeanKeyword{set_option} maxHeartbeats 800000
\LeanKeyword{open} Fintype Set Real Ideal Polynomial
\LeanKeyword{open} \LeanKeyword{scoped} BigOperators
\LeanKeyword{open} Fintype Set Real Ideal Polynomial
\LeanKeyword{open} \LeanKeyword{scoped} BigOperators
\LeanKeyword{theorem} exercise_4_3_1 \{R : \LeanKeyword{Type}*\}
    [CommRing R] (a : R) :
    ∃ I : Ideal R, \{x : R | x*a=0\} = I := \LeanKeyword{by}
  \color{leanDelimiter}/- <feedback>
  \color{leanGoal}-- proof state:
  \color{leanGoal}R : Type u_1
  \color{leanGoal}inst✝ : CommRing R
  \color{leanGoal}a : R
  \color{leanGoal}⊢ ∃ I, \{x | x * a = 0\} = ↑I
  \color{leanDelimiter}</feedback> -/
  \LeanKeyword{have} h_main : ∃ (I : Ideal R),
      \{x : R | x*a=0\} = I := \LeanKeyword{by}
    \color{leanDelimiter}/- <feedback>
    \color{leanGoal}-- proof state:
    \color{leanGoal}⊢ ∃ I, \{x | x * a = 0\} = ↑I
    \color{leanDelimiter}</feedback> -/
    \LeanKeyword{refine'} ⟨\{x : R | x * a = 0\}, _⟩
    \color{leanDelimiter}/- <feedback>
    \color{leanError}-- type: error, msg: type mismatch
      \color{leanError}\{x | x * a = 0\}
    \color{leanError}has type
      \color{leanError}Set R : Type u_1
    \color{leanError}but is expected to have type
      \color{leanError}Ideal R : Type u_1
    \color{leanDelimiter}</feedback> -/
\end{Verbatim}
\end{minipage}
\caption{\textbf{Invalid proof: error detected, message paraphrased.} For ProofNet \texttt{exercise\_4\_3\_1}, Lean rejects the \texttt{refine'} step because a set is supplied where an \texttt{Ideal R} is expected, and the model predicts invalid with an error block at the same tactic. The two goal-state blocks match exactly. The error blocks differ: Lean reports an \emph{application type mismatch} for the elaborated application \texttt{Exists.intro \{x | x * a = 0\}} and lists the offending argument, whereas the model predicts a plain \emph{type mismatch} that names only the set. The stated conflict, \texttt{Set R} versus \texttt{Ideal R}, is the same. Block F1 and anchored block F1 are $0.667$, token F1 is $0.929$, and error-presence agreement is $1$. Both complete saved traces are shown, including imports and options; only whitespace is reflowed. Scores use the original traces.}
\label{fig:feedback-paraphrase}
\end{figure}

\begin{figure}[!ht]
\centering
\begin{minipage}[t]{0.49\textwidth}
\vspace{0pt}
\centering\small\textbf{Lean annotation}
\par\smallskip
\raggedright\fontsize{8}{9.2}\selectfont
\begin{Verbatim}[commandchars=\\\{\},breaklines,breakanywhere=false,breaksymbolleft={},breaksymbolright={}]
\LeanKeyword{import} Mathlib
\LeanKeyword{set_option} maxHeartbeats 800000
\LeanKeyword{open} Fintype Subgroup Set Polynomial Ideal
\LeanKeyword{open} \LeanKeyword{scoped} BigOperators
\LeanKeyword{open} Fintype Subgroup Set Polynomial Ideal
\LeanKeyword{open} \LeanKeyword{scoped} BigOperators
\LeanKeyword{theorem} exercise_1_1_22a
    \{G : \LeanKeyword{Type}*\} [Group G] (x g : G) :
    orderOf x = orderOf (g⁻¹ * x * g) := \LeanKeyword{by}
  \color{leanDelimiter}/- <feedback>
  \color{leanGoal}-- proof state:
  \color{leanGoal}G : Type u_1
  \color{leanGoal}inst✝ : Group G
  \color{leanGoal}x : G
  \color{leanGoal}g : G
  \color{leanGoal}⊢ orderOf x = orderOf (g⁻¹ * x * g)
  \color{leanDelimiter}</feedback> -/
  \LeanKeyword{have} h_main :
      orderOf x = orderOf (g⁻¹ * x * g) := \LeanKeyword{by}
    \color{leanDelimiter}/- <feedback>
    \color{leanGoal}-- proof state:
    \color{leanGoal}⊢ orderOf x = orderOf (g⁻¹ * x * g)
    \color{leanDelimiter}</feedback> -/
    \LeanKeyword{apply} orderOf_conj
    \color{leanDelimiter}/- <feedback>
    \color{leanError}-- type: error, msg: unknown identifier
      \color{leanError}'orderOf_conj'
    \color{leanDelimiter}</feedback> -/
\end{Verbatim}
\end{minipage}\hfill
\begin{minipage}[t]{0.49\textwidth}
\vspace{0pt}
\centering\small\textbf{Model prediction}
\par\smallskip
\raggedright\fontsize{8}{9.2}\selectfont
\begin{Verbatim}[commandchars=\\\{\},breaklines,breakanywhere=false,breaksymbolleft={},breaksymbolright={}]
\LeanKeyword{import} Mathlib
\LeanKeyword{set_option} maxHeartbeats 800000
\LeanKeyword{open} Fintype Subgroup Set Polynomial Ideal
\LeanKeyword{open} \LeanKeyword{scoped} BigOperators
\LeanKeyword{open} Fintype Subgroup Set Polynomial Ideal
\LeanKeyword{open} \LeanKeyword{scoped} BigOperators
\LeanKeyword{theorem} exercise_1_1_22a
    \{G : \LeanKeyword{Type}*\} [Group G] (x g : G) :
    orderOf x = orderOf (g⁻¹ * x * g) := \LeanKeyword{by}
  \color{leanDelimiter}/- <feedback>
  \color{leanGoal}-- proof state:
  \color{leanGoal}G : Type u_1
  \color{leanGoal}inst✝ : Group G
  \color{leanGoal}x : G
  \color{leanGoal}g : G
  \color{leanGoal}⊢ orderOf x = orderOf (g⁻¹ * x * g)
  \color{leanDelimiter}</feedback> -/
  \LeanKeyword{have} h_main :
      orderOf x = orderOf (g⁻¹ * x * g) := \LeanKeyword{by}
    \color{leanDelimiter}/- <feedback>
    \color{leanGoal}-- proof state:
    \color{leanGoal}⊢ orderOf x = orderOf (g⁻¹ * x * g)
    \color{leanDelimiter}</feedback> -/
    \LeanKeyword{apply} orderOf_conj
    <;> \LeanKeyword{simp} [mul_assoc]
    <;> \LeanKeyword{aesop}

  \color{leanDelimiter}/- <feedback>
  \color{leanGoal}-- proof state:
  \color{leanGoal}h_main : orderOf x = orderOf (g⁻¹ * x * g)
  \color{leanGoal}⊢ orderOf x = orderOf (g⁻¹ * x * g)
  \color{leanDelimiter}</feedback> -/
  \LeanKeyword{apply} h_main
\end{Verbatim}
\end{minipage}
\caption{\textbf{Invalid proof: a missed compiler error.} For ProofNet \texttt{exercise\_1\_1\_22a}, Lean reports that \texttt{orderOf\_conj} is unknown, but the model continues annotating and predicts valid. Block F1 is $0.667$, anchored block F1 is $0.667$, token F1 is $0.745$, and error-presence agreement is $0$. Both complete saved traces are shown; the Lean annotation ends at the first error by construction. Only whitespace is reflowed, and scores use the original traces.}
\label{fig:feedback-invalid}
\end{figure}

\end{document}